\documentclass[10pt,twocolumn,letterpaper]{article}

\usepackage{cvpr}              
\usepackage{xcolor}
\usepackage{comment}

\definecolor{cvprblue}{rgb}{0.21,0.49,0.74}
\usepackage[pagebackref,breaklinks,colorlinks,citecolor=cvprblue]{hyperref}

\def\paperID{*****} 
\def\confName{CVPR}
\def\confYear{2024}

\title{Simple In-place Data Augmentation for Surveillance Object Detection}

\begin{document}
\author{Munkh-Erdene Otgonbold$^{1,3}$ 
\hspace{0.1cm} Ganzorig Batnasan$^{1}$ 
\hspace{0.1cm} Munkhjargal Gochoo$^{1,2}$ 
\\
\\
$^{1}$ Department of Computer Science and Software Engineering, United Arab Emirates University, UAE\\
$^{2}$ Emirates Center for Mobility Research, United Arab Emirates University, UAE\\
$^{3}$ Department of Electronics, Mongolian University of Science and Technology, Mongolia\\
\small 
\url{omunkuush@uaeu.ac.ae}, \hspace{1mm} 
\url{gbatnasan@uaeu.ac.ae}, \hspace{1mm} 
\url{mgochoo@uaeu.ac.ae}, \hspace{1mm}\\
}

\maketitle
\begin{abstract}
Motivated by the need to improve model performance in traffic monitoring tasks with limited labeled samples, we propose a straightforward augmentation technique tailored for object detection datasets, specifically designed for stationary camera-based applications. Our approach focuses on placing objects in the same positions as the originals to ensure its effectiveness. By applying in-place augmentation on objects from the same camera input image, we address the challenge of overlapping with original and previously selected objects. Through extensive testing on two traffic monitoring datasets, we illustrate the efficacy of our augmentation strategy in improving model performance, particularly in scenarios with limited labeled samples and imbalanced class distributions. Notably, our method achieves comparable performance to models trained on the entire dataset while utilizing only 8.5 percent of the original data. Moreover, we report significant improvements, with  mAP@.5 increasing from 0.4798 to 0.5025, and the mAP@.5:.95 rising from 0.29 to 0.3138 on the FishEye8K dataset. These results highlight the potential of our augmentation approach in enhancing object detection models for traffic monitoring applications.
\end{abstract}
\section{Introduction}

Traffic cameras are extensively utilized for monitoring traffic conditions, particularly in areas surrounding intersections. Vision algorithms for cameras have been developed to automate a range of tasks, such as detecting and tracking vehicles and pedestrians \cite{Fedorov2019}, as well as re-identification \cite{Khan2019}. Object detection, a fundamental task of computer vision, plays a crucial role in understanding visual scenes. This process includes pinpointing the location of the objects and outlining precise bounding boxes around them. Essentially, it enables the computer to recognize and categorize different items present in a given image. 
\begin{figure}[t]
  \centering
   \includegraphics[width=1\linewidth]{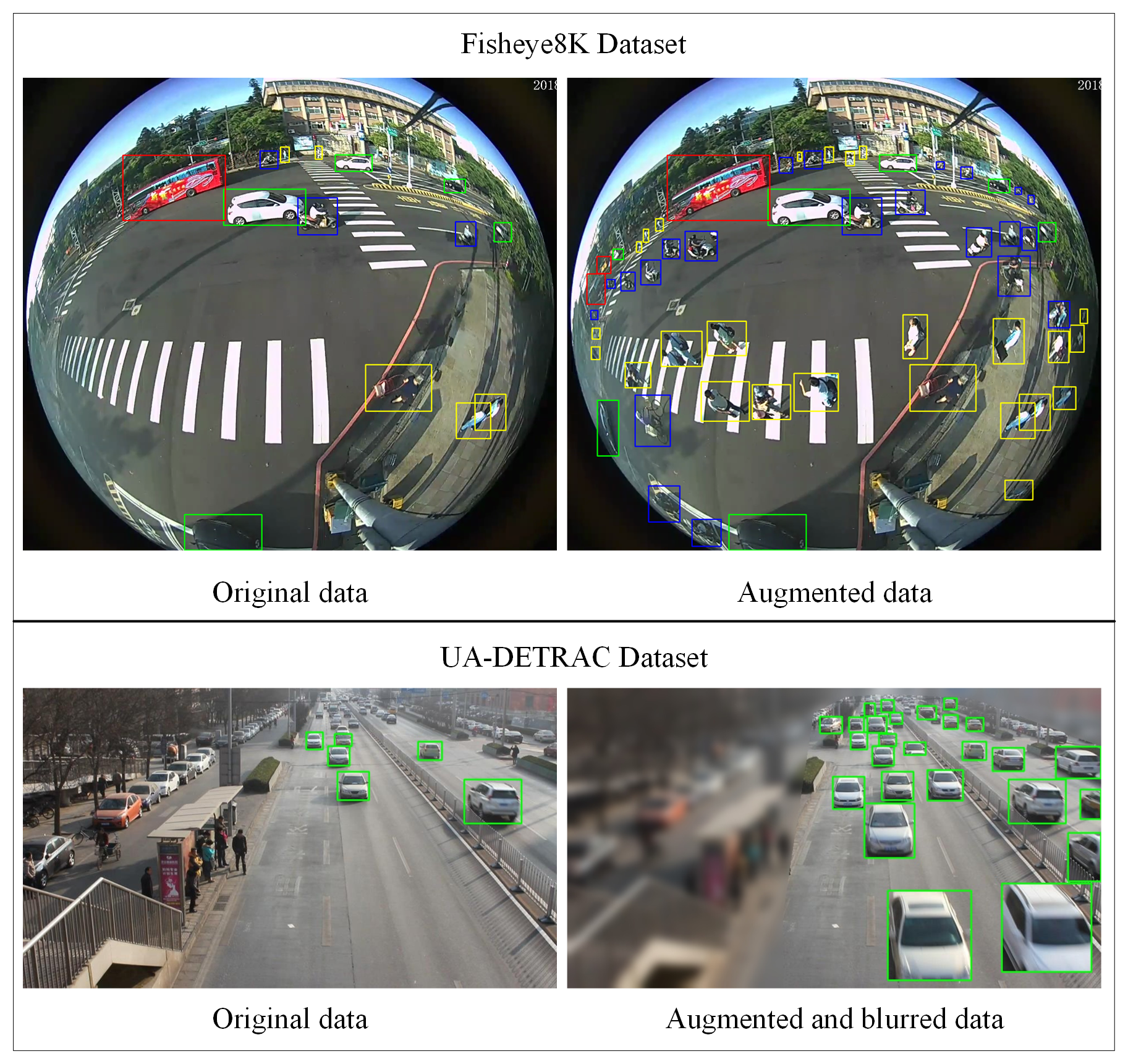}
   \caption{The comparison between the original sample images and augmented images of FishEye8K and UA-DETRAC datasets. Both augmented samples include a comparatively larger number of objects due to the in-place augmentation. In contrast, the UA-DETRAC sample has blurred areas, which are the regions of non-interest determined by subtracting the polygonal area of the bounding boxes of all the objects.}
   \label{fig:comparison}
\end{figure}
The field of computer vision has experienced significant advancements across different areas, extending beyond traffic monitoring. Applications like face recognition \cite{Ranjan2019}, robotic grasping \cite{Kumra2017}, and human interaction \cite{Gkioxari2018} have all benefited from these developments, largely driven by the rapid progress of deep convolutional neural networks (CNNs) \cite{He2016}, \cite{Huang2017}, \cite{Simonyan2014VeryDC}, \cite{Szegedy2017}. While CNN architectures have shown exceptional performance, their effectiveness heavily relies on the availability of extensive sets of accurately labeled training data. For object detection models, the need for data augmentation is more crucial as collecting labeled data for detection is more costly and common detection datasets have many fewer examples than image classification datasets. 
A more advanced method for data augmentation involves utilizing segmentation annotations, which can be acquired either through manual efforts or generated by an automated segmentation system. This technique involves generating new images by positioning objects at different locations within pre-existing scenes \cite{Dwibedi2017}, \cite{Gupta2016}, \cite{Georgakis2017}. Though it doesn't attain flawless photorealism, the approach of employing random placements has demonstrated unexpected effectiveness for object instance detection \cite{Dwibedi2017}, which is a fine grained detection.  In contrast, object detection concentrates on identifying instances of objects from a specified category. By placing training objects at unrealistic positions, implicitly modeling context becomes difficult and detection accuracy drops substantially. 

Achieving balance in the number of instances per class within a traffic monitoring dataset presents a formidable challenge, primarily stemming from the extensive diversity observed across various regions. The dataset encompasses a wide range of geographical locations, each characterized by unique traffic patterns, infrastructure layouts, and behavioral norms. Road objects from specific classes tend to appear more in specific region (e.g., bikes in warm countries, less pedestrians in countries with extreme weather ). 
Consequently, ensuring equitable representation of these diverse elements necessitates meticulous consideration and strategic planning. 
Balancing the classes becomes particularly intricate as disparities in traffic volume, road conditions, and cultural practices manifest differently across regions, complicating efforts to maintain proportional class distribution. 
Therefore, meticulous attention to detail and adaptive methodologies are imperative in addressing these inherent complexities and achieving a harmonious balance in the dataset's class distribution.

We propose to tackle the issue by applying in-place augmentation on the same position from the same camera. This increases the diversity in the locations of traffic objects while ensuring that those objects appear from correct angle. In Figure \ref{fig:comparison}, we illustrate sample augmentation generated by our proposed method. The figure contrasts an original image without any augmentation with an augmented image sample, amplified by a factor of 20X, sourced from the Fisheye8K \cite{fisheye8K} and UA-DETRAC \cite{ua_detrac} datasets.

To enhance the object's impact on the model, we increased the number of objects within the image rather than augmenting the number of images. We selected and placed objects from the same camera input image that could be placed without overlapping with original and previously selected objects. 
We show with extensive tests on two traffic monitoring datasets that our augmentation approach can be used for improving model performance when few labeled samples are available.

\section{Related Works}
Zoph et al. \cite{Barret_Zoph} delve into the enhancement of generalization performance for detection models through the exploration of learned, specialized data augmentation policies. With meticulous curation, they assembled subsets of images from the COCO dataset \cite{coco_dataset}, ranging in size from 5000 to 23000 images. The researchers observed a notable improvement in detection accuracy of more than +2.3 mAP across different ResNet backbones, resulting in mAP values ranging from 39.0 to 42.1. Furthermore, when applied to a distinct detection model featuring an AmoebaNet-D backbone \cite{AmoebaNet_D}, the method achieved a remarkable increase of +1.5\% mAP, attaining a state-of-the-art accuracy of 50.7 mAP.  

Ghiasi et al. \cite{copypaste} explores the efficacy of Copy-Paste augmentation for instance segmentation, revealing that random object pasting yields significant performance improvements over previous methods focusing on contextual modeling. Moreover, they demonstrate that integrating Copy-Paste with semi-supervised techniques that utilize additional data via pseudo-labeling, such as self-training, yields notable enhancements. Their approach achieves a mask AP of 49.1 and a box AP of 57.3 on COCO instance segmentation, surpassing the previous state-of-the-art by +0.6 mask AP and +1.5 box AP.

Dvornik et al. \cite{nikita_dvornik} proposed a data augmentation method consisting of two main steps: first, utilizing bounding box annotations to model visual context and train a CNN to predict object presence or absence; second, employing the trained context model to generate new object locations. Their approach, applied to a subset of the Pascal VOC’12 dataset \cite{pascal_voc_2012}, involved training a single multiple-category object detector with significantly more labeled data, resulting in a 1.3\% average improvement over baseline across various categories

\begin{figure*}[t]
  \centering
   \includegraphics[width=1\linewidth]{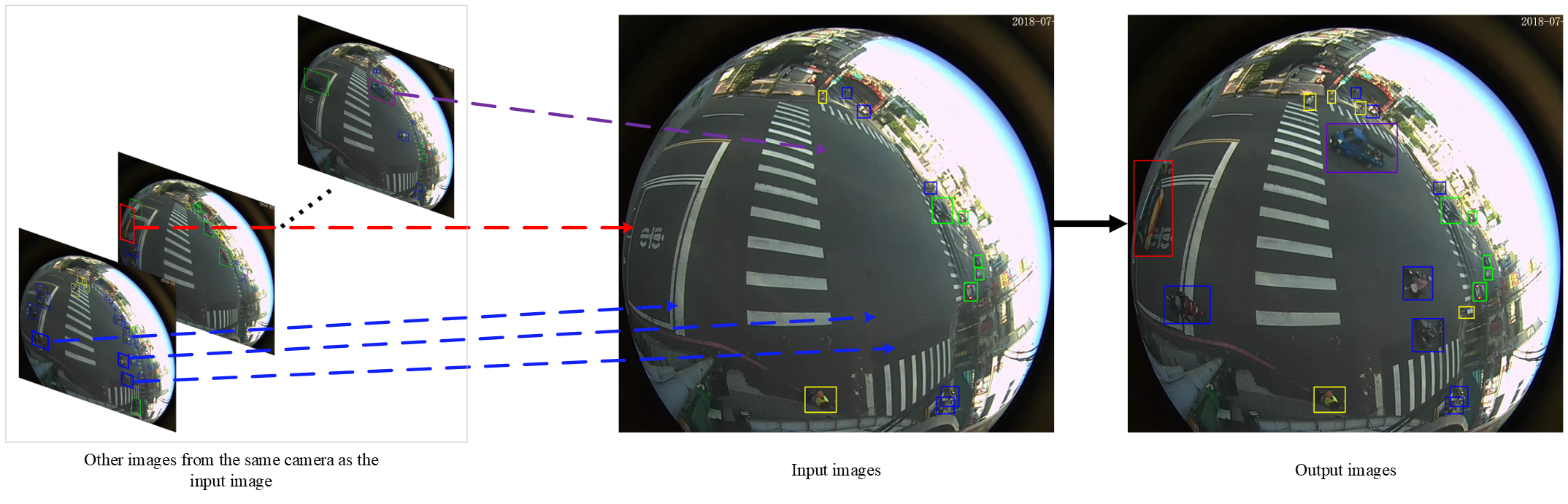}
   \caption{In-place object augmentation method on Fisheye8k dataset. An augmented output sample image has multiple objects that appear on the other frames of the same surveillance camera video.}
   \label{fig:aug}
\end{figure*}

Cubuk et al. \cite{ekin_cubuk} propose a simplified approach to automated augmentation strategies, eliminating the need for a separate search phase. They apply their method across CIFAR-10/100, SVHN, ImageNet, and COCO datasets \cite{coco_dataset}. Notably, using EfficientNet-B7, they achieve a 1.0\% increase in accuracy over baseline augmentation and a 0.6\% improvement over AutoAugment on the ImageNet dataset. 

Behpour et al. \cite{behpour_sima} offers a game-theoretic perspective on data augmentation in object detection, seeking optimal adversarial perturbations of ground truth data to enhance test-time performance. They demonstrate significant improvements of approximately 16\%, 5\%, and 2\% respectively on the ImageNet\cite{imageNet_1k}, Pascal VOC \cite{pascal_voc_2007}, and MS-COCO \cite{coco_dataset} object detection tasks compared to leading data augmentation methods. 

Kisantal et al. \cite{kisantal_mate} tackle the performance gap in object detection between small and large objects by employing Mask-RCNN on the MS COCO dataset \cite{coco_dataset}. Their approach involves oversampling images containing small objects and augmenting them through the repeated copy-pasting of small objects. This method leads to a significant 9.7\% relative enhancement in instance segmentation and a 7.1\% improvement in object detection of small objects compared to the current state-of-the-art performance on MS COCO \cite{coco_dataset}. 

Shao et al. \cite{shao_shitong} address the limitations of SRe2L's "local-match-global" matching method by introducing "generalized matching" through G-VBSM. Their approach surpasses state-of-the-art methods by 3.9\%, 6.5\%, and 10.1\% on CIFAR-100 \cite{cifar10_100}, Tiny-ImageNet \cite{tiny_imageNet}, and ImageNet-1k \cite{imageNet_1k}, respectively, demonstrating superior performance across small and large-scale datasets. 

These methods have achieved favorable results through the integration of traditional augmentation methods with other techniques. While effective in enhancing results, the augmentation method poses significant computational demands due to the increased volume of data. However, these augmentation methods often struggle to preserve the realism of the images.

\section{Datasets}
\begin{figure}[h!]
  \centering
   \includegraphics[width=1\linewidth]{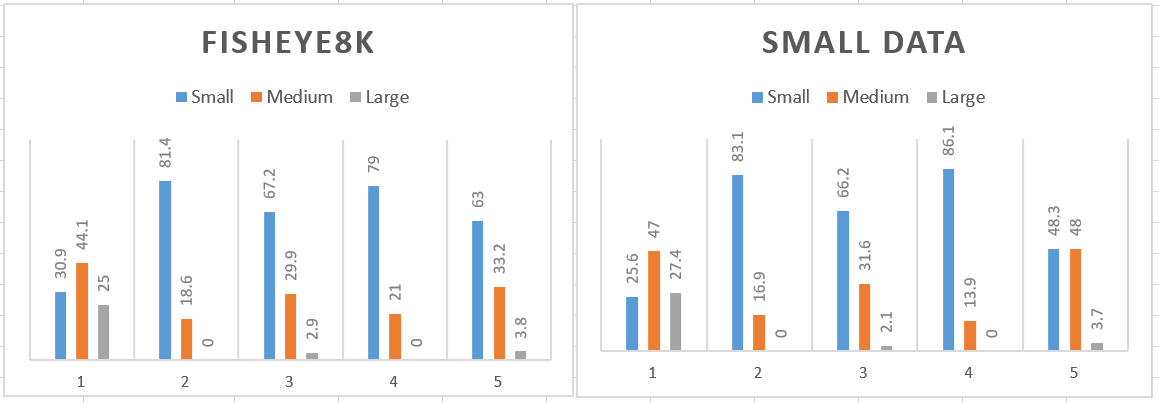}
   \caption{Number of objects of (a) Full dataset and (b) Sampled small dataset that is 8.5 percent of the full dataset.}
   \label{fig:bar_graph}
\end{figure}

The Fisheye8K \cite{fisheye8K} dataset contains 8000 images from 18 different cameras with 157K bounding boxes for five object classes. Of this, 5287 images of 14 cameras in the train set are divided into 2712 images of 4 cameras in the test set. We have selected 450 images, a small subset from the Fisheye8K \cite{fisheye8K} dataset's train set, which is only 8.5\% of the full train set. Figure \ref{fig:bar_graph} shows a comparison bar graph of small, medium, and large objects in the Fisheye8K dataset's trainset and selected small set. The proportion of small, medium, and large objects in the 2 datasets is almost the same, and it is slightly different for the last 5th class. 

UA-DETRAC \cite{ua_detrac} dataset comprises 100 videos which is divided into 83k images from 60 sequences for training set and  56k images from 40 sequences for testing set. We have selected 7140 images small set from UA DETRAC \cite{ua_detrac} dataset's train set. which is 8.5\% of the full train set. 
As for 8.5\%, after selecting a dataset that fully represents the Fisheye8K \cite{fisheye8K} dataset and achieving good results, the UA DETRAC \cite{ua_detrac} dataset also took a smaller portion from the larger dataset with the same percentage.

\begin{figure*}[t]
  \centering
   \includegraphics[width=1\linewidth]{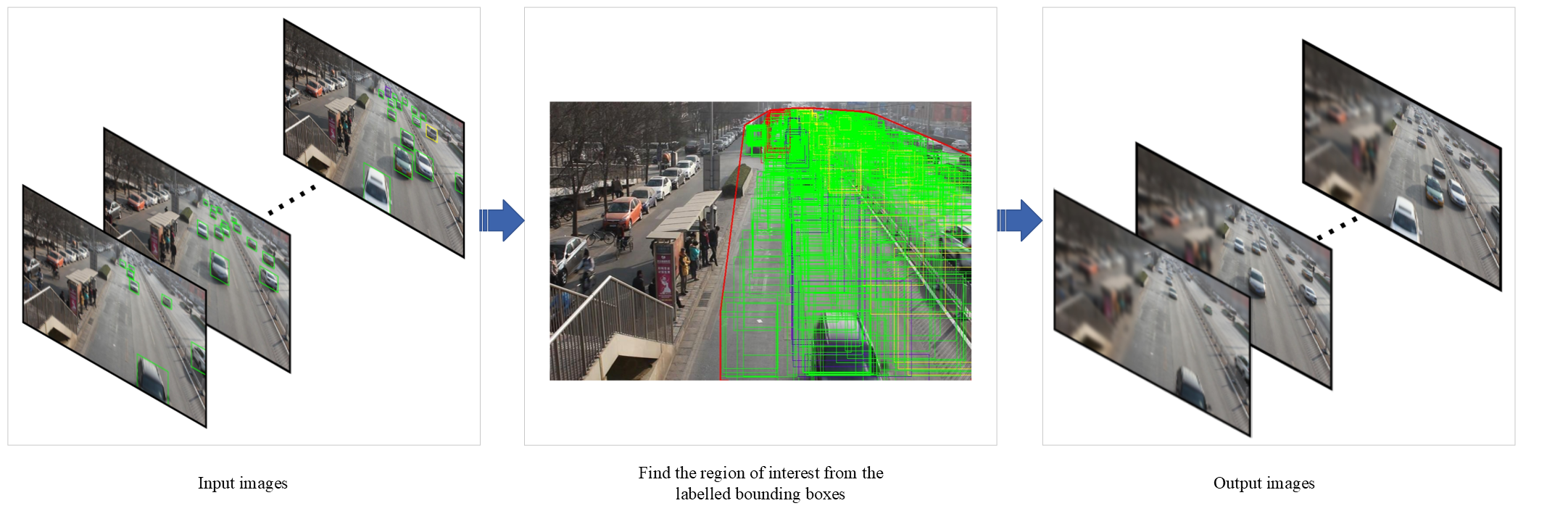}
   \caption{Determining the region of interest, the polygonal area drawn in red, from all object labels in the specific camera video. }
   \label{fig:region_blur}
\end{figure*}

\section{Augmentation}
We introduce a stationary camera-based object augmentation method, where Figure \ref{fig:aug} shows an example of our proposed augmentation process. When performing a data augmentation, to improve the effect of the object on the model, we augmented the number of objects in the image, not the number of images. On the contrary, we tried to reach and outperform the accuracy of the full dataset with as small as possible portion of the full set. In doing so, from all objects captured in the same camera video as the input image, those that can be positioned in the original location in the image without overlapping with others are chosen and placed accordingly. We opted for 3, 10, and 20 objects as values for augmentation. The model performances trained on the augmented data are presented in the results section under the names: "augmented 3X," "augmented 10X," "augmented 20X," and "Assembled". For Augmented 3X, 10X, and 20X, a selection of 3, 10, and 20 objects from each class, taken by the same camera as the input image, were placed onto the input image. These objects were then positioned onto the input image, ensuring they did not overlap with any previously placed objects. During the object selection process, images had to belong to either the day or night category, consistent with the input image. For Assembled, from the results of the above datasets: small data, augmented 3X, augmented 10X, augmented 20X, the objects of each class's objects with the best results were combined to form the data that gives the best results.
Table \ref{tb:objects} shows the increase in objects for each class in a small dataset from the Fisheye8K \cite{fisheye8K} dataset.

\begin{table}[h!]
\scriptsize
\centering
\begin{tabular}{|l|c|c|c|c|c|}
\hline
           & \multicolumn{1}{l|}{Small Data} & \multicolumn{1}{l|}{3X} & \multicolumn{1}{l|}{10X} & \multicolumn{1}{l|}{20X} & \multicolumn{1}{l|}{Assembled} \\ \hline
Bus        & 219                           & 531                              & 441                                 & 428                                 & 459                           \\ \hline
Bike       & 5060                          & 6375                             & 9296                                & 12618                               & 9296                          \\ \hline
Car        & 3720                          & 5022                             & 7298                                & 8038                                & 3720                          \\ \hline
Pedestrian & 812                           & 1765                             & 3434                                & 4298                                & 812                           \\ \hline
Truck      & 325                           & 851                              & 773                                 & 692                                 & 325                           \\ \hline
\end{tabular}
\caption{Number of original and augmented objects in a small dataset from the Fisheye8K dataset.}
\label{tb:objects}
\end{table}

For the UA DETRAC \cite{ua_detrac} dataset, only objects in traffic are labeled and objects belonging to the same class standing on the side of the road are not labeled, which makes it difficult for any model to learn the data in that dataset. To solve this problem, we detect the active traffic areas of each camera and blurred the rest of the areas, it makes the data easier to learn for object detection. Figure \ref{fig:region_blur} shows how to blur the image as an example. 
The augmentation in object counts for each class within a small set of the UA-DETRAC dataset \cite{ua_detrac} is presented in Table \ref{tb:num_objects_detrac}.

\begin{table}[h!]
\scriptsize
\centering
\begin{tabular}{|l|c|c|c|c|c|}
\hline
       & Small Data & 3X    & 10X    & 20X    & Assembled \\ \hline
Others & 321       & 3489  & 2450   & 2235   & 2449      \\ \hline
Car    & 44256     & 65556 & 106479 & 129140 & 44256     \\ \hline
Van    & 4756      & 20291 & 28117  & 25727  & 25071     \\ \hline
Bus    & 3221      & 8656  & 7257   & 6814   & 6570      \\ \hline
\end{tabular}
\caption{Number of original and augmented objects in a small dataset from the UA-DETRAC dataset.}
\label{tb:num_objects_detrac}
\end{table}

\section{Results}
We trained the YOLOv7-E6E \cite{yolov7} model by augmenting two extensive traffic datasets, Fisheye8K \cite{fisheye8K} and UA DETRAC \cite{ua_detrac}, using our custom augmentation technique, resulting in multiple outcomes. Depending on the primary image size, we set the input size to 1280x1280 for the Fisheye8K \cite{fisheye8K} dataset and 640x640 for the UA-DETRAC \cite{ua_detrac} dataset. During all training procedures, a pre-trained model trained on the COCO dataset \cite{coco_dataset} was utilized, with both the IoU threshold and confidence threshold set to 0.5 during evaluation. For Fisheye8K \cite{fisheye8K} dataset's result, All evaluations were made on the original validation set. For UA-DETRAC \cite{ua_detrac} dataset's results, the model trained on small dataset was evaluated on the original validation set and other models trained on blurred datasets were evaluated on the blurred validation set.

\subsection{Fisheye8K}

\begin{figure}[t]
  \centering
   \includegraphics[width=1\linewidth]{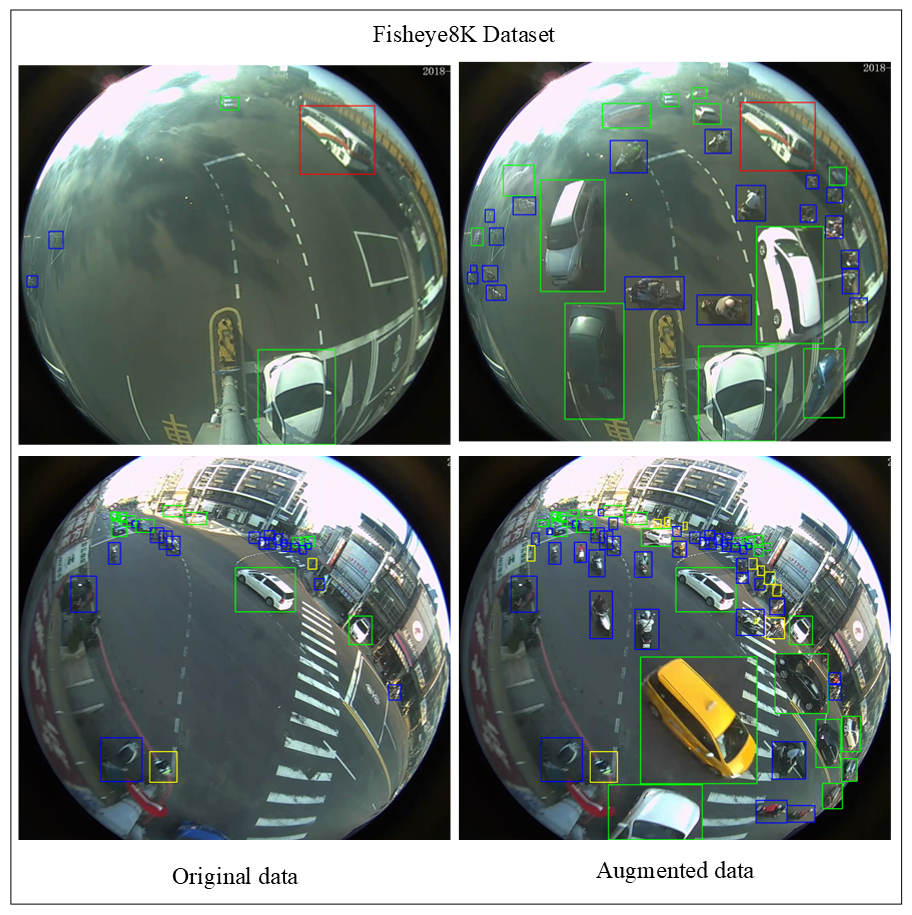}
   \caption{The comparison between original image and augmented image in Fisheye8K dataset.}
   \label{fig:fisheye8K_example}
\end{figure}

Figure \ref{fig:fisheye8K_example} shows the original and augmented samples in Fisheye8K dataset \cite{fisheye8K}. The results depicted in Table \ref{tb:smalldata} illustrate the performance of the YOLOv7-E6E \cite{yolov7} model trained on a modest dataset without any augmentation. Notably, the model attained its peak mAP@.5 of 0.4267 in detecting cars, while registering its lowest mAP@.5 of 0.1449 in pedestrian detection. Notably, it showed great performance in detecting buses, with a mAP@.5 of 0.4137. Overall, the model demonstrated a mAP@.5 of 0.4798 and an F1-score of 0.62.

\begin{table}[h!]
\centering
\scriptsize
\begin{tabular}{llllllllll}
\cline{1-6}
\multicolumn{6}{|c|}{Small Data}      \\ \cline{1-6}
\multicolumn{1}{|l|}{}             & \multicolumn{1}{l|}{Precision}       & \multicolumn{1}{l|}{Recall}          & \multicolumn{1}{l|}{mAP@.5}           & \multicolumn{1}{l|}{mAP@.5:.95}        & \multicolumn{1}{l|}{f1-score}  \\ \cline{1-6}
\multicolumn{1}{|l|}{Bus}          & \multicolumn{1}{l|}{0.9221}          & \multicolumn{1}{l|}{0.5602}          & \multicolumn{1}{l|}{0.5542}          & \multicolumn{1}{l|}{0.4137}        & \multicolumn{1}{l|}{0.697}      \\ \cline{1-6}
\multicolumn{1}{|l|}{Bike}         & \multicolumn{1}{l|}{0.7734}          & \multicolumn{1}{l|}{0.5142}          & \multicolumn{1}{l|}{0.4829}          & \multicolumn{1}{l|}{0.2344}        & \multicolumn{1}{l|}{0.6177}     \\ \cline{1-6}
\multicolumn{1}{|l|}{Car}          & \multicolumn{1}{l|}{0.8238}          & \multicolumn{1}{l|}{0.6756}          & \multicolumn{1}{l|}{0.6589}          & \multicolumn{1}{l|}{0.4267}        & \multicolumn{1}{l|}{0.7424}  \\ \cline{1-6}
\multicolumn{1}{|l|}{Pedestrian}   & \multicolumn{1}{l|}{0.7592}          & \multicolumn{1}{l|}{0.2993}          & \multicolumn{1}{l|}{0.2764}          & \multicolumn{1}{l|}{0.1449}        & \multicolumn{1}{l|}{0.4293}  \\ \cline{1-6}
\multicolumn{1}{|l|}{Truck}        & \multicolumn{1}{l|}{0.7872}          & \multicolumn{1}{l|}{0.5029}          & \multicolumn{1}{l|}{0.4265}          & \multicolumn{1}{l|}{0.2303}        & \multicolumn{1}{l|}{0.6137} \\ \cline{1-6}
\multicolumn{1}{|l|}{All} & \multicolumn{1}{l|}{0.8132} & \multicolumn{1}{l|}{0.5104} & \multicolumn{1}{l|}{0.4798} & \multicolumn{1}{l|}{0.29} & \multicolumn{1}{l|}{0.62}           \\ \cline{1-6}
\end{tabular}
\caption{Result of YOLOv7-e6e model on the small dataset.}
\label{tb:smalldata}
\end{table}

The data provided in Table \ref{tb:smalldata3x} indicates that the model excelled in detecting buses, achieving a mAP@.5 of 0.4811 and F1-score of 0.7249. Conversely, its performance was least effective in detecting pedestrians, with a mAP@.5 of 0.1224 and an F1-score of 0.3667. Additionally, the model demonstrated great performance in detecting cars, achieving a mAP@.5 of 0.4069 and an F1-score of 0.7212. Overall, the model yielded a mAP@.5 of 0.3054 and an F1-score of 0.5859.

\begin{table}[h!]
\centering
\scriptsize
\begin{tabular}{llllllllll}
\cline{1-6}
\multicolumn{6}{|c|}{Small Data 3X}  \\ \cline{1-6}
\multicolumn{1}{|l|}{}             & \multicolumn{1}{l|}{Precision}       & \multicolumn{1}{l|}{Recall}          & \multicolumn{1}{l|}{mAP@.5}           & \multicolumn{1}{l|}{mAP@.5:.95}                                  & \multicolumn{1}{l|}{f1-score}     \\ \cline{1-6}
\multicolumn{1}{|l|}{Bus}          & \multicolumn{1}{l|}{0.8037}          & \multicolumn{1}{l|}{0.6602}          & \multicolumn{1}{l|}{0.641}           & \multicolumn{1}{l|}{0.4811}                                  & \multicolumn{1}{l|}{0.7249}  \\ \cline{1-6}
\multicolumn{1}{|l|}{Bike}         & \multicolumn{1}{l|}{0.7705}          & \multicolumn{1}{l|}{0.4909}          & \multicolumn{1}{l|}{0.4625}          & \multicolumn{1}{l|}{0.2243}                                  & \multicolumn{1}{l|}{0.5997}                         \\ \cline{1-6}
\multicolumn{1}{|l|}{Car}          & \multicolumn{1}{l|}{0.8338}          & \multicolumn{1}{l|}{0.6354}          & \multicolumn{1}{l|}{0.6237}          & \multicolumn{1}{l|}{0.4069}                                  & \multicolumn{1}{l|}{0.7212}                        \\ \cline{1-6}
\multicolumn{1}{|l|}{Pedestrian}   & \multicolumn{1}{l|}{0.7882}          & \multicolumn{1}{l|}{0.2389}          & \multicolumn{1}{l|}{0.2252}          & \multicolumn{1}{l|}{0.1224}                                  & \multicolumn{1}{l|}{0.3667}                       \\ \cline{1-6}
\multicolumn{1}{|l|}{Truck}        & \multicolumn{1}{l|}{0.8177}          & \multicolumn{1}{l|}{0.3778}          & \multicolumn{1}{l|}{0.3732}          & \multicolumn{1}{l|}{0.2924}                                  & \multicolumn{1}{l|}{0.5168}                       \\ \cline{1-6}
\multicolumn{1}{|l|}{All} & \multicolumn{1}{l|}{0.8028} & \multicolumn{1}{l|}{0.4806} & \multicolumn{1}{l|}{0.4651} & \multicolumn{1}{l|}{0.3054} & \multicolumn{1}{l|}{0.5859}             \\ \cline{1-6}
\end{tabular}
\caption{Result of YOLOv7-e6e model on the small dataset with a 3X augmentation.}
\label{tb:smalldata3x}
\end{table}

In Table \ref{tb:smalldata10x}, the model demonstrated notable performance in detecting cars, achieving a mAP@.5 of 0.6403 and an F1-score of 0.727. However, its performance was comparatively lower in detecting pedestrians, with a mAP@.5 of 0.262 and an F1-score of 0.3972. Additionally, the model showed satisfactory performance in detecting buses, achieving a mAP@.5 of 0.5727 and an F1-score of 0.7089. Overall, the model yielded a mAP@.5 of 0.4931 and an F1-score of 0.6141.

\begin{table}[h!]
\centering
\scriptsize
\begin{tabular}{llllllllll}
\cline{1-6}
\multicolumn{6}{|c|}{Small Data 10X}    \\ \cline{1-6}
\multicolumn{1}{|l|}{}             & \multicolumn{1}{l|}{Precision}       & \multicolumn{1}{l|}{Recall}                                  & \multicolumn{1}{l|}{mAP@.5}                                   & \multicolumn{1}{l|}{mAP@.5:.95}          & \multicolumn{1}{l|}{f1-score}    \\ \cline{1-6}
\multicolumn{1}{|l|}{Bus}          & \multicolumn{1}{l|}{0.902}           & \multicolumn{1}{l|}{0.5839}                                  & \multicolumn{1}{l|}{0.5727}                                  & \multicolumn{1}{l|}{0.4029}          & \multicolumn{1}{l|}{0.7089}     \\ \cline{1-6}
\multicolumn{1}{|l|}{Bike}         & \multicolumn{1}{l|}{0.6893}          & \multicolumn{1}{l|}{0.5697}                                  & \multicolumn{1}{l|}{0.5152}                                  & \multicolumn{1}{l|}{0.2425}          & \multicolumn{1}{l|}{0.6238}   \\ \cline{1-6}
\multicolumn{1}{|l|}{Car}          & \multicolumn{1}{l|}{0.8115}          & \multicolumn{1}{l|}{0.6584}                                  & \multicolumn{1}{l|}{0.6403}                                  & \multicolumn{1}{l|}{0.4094}          & \multicolumn{1}{l|}{0.727}   \\ \cline{1-6}
\multicolumn{1}{|l|}{Pedestrian}   & \multicolumn{1}{l|}{0.5962}          & \multicolumn{1}{l|}{0.2978}                                  & \multicolumn{1}{l|}{0.262}                                   & \multicolumn{1}{l|}{0.1376}          & \multicolumn{1}{l|}{0.3972}   \\ \cline{1-6}
\multicolumn{1}{|l|}{Truck}        & \multicolumn{1}{l|}{0.7765}          & \multicolumn{1}{l|}{0.5071}                                  & \multicolumn{1}{l|}{0.4753}                                  & \multicolumn{1}{l|}{0.2906}          & \multicolumn{1}{l|}{0.6135}    \\ \cline{1-6}
\multicolumn{1}{|l|}{All} & \multicolumn{1}{l|}{0.7551} & \multicolumn{1}{l|}{0.5234} & \multicolumn{1}{l|}{0.4931} & \multicolumn{1}{l|}{0.2966} & \multicolumn{1}{l|}{0.6141}      \\ \cline{1-6}

\end{tabular}
\caption{Result of YOLOv7-e6e model on the small dataset with a 10X augmentation.}
\label{tb:smalldata10x}
\end{table}

Table \ref{tb:smalldata20x} illustrates the model's better performance in car detection, achieving a mAP@.5 of 0.6623 and an associated F1-score of 0.7288. However, its performance was less satisfactory in pedestrian detection, recording its lowest mAP@.5 at 0.2113, with an F1-score of 0.3486. Furthermore, the model exhibited commendable performance in detecting buses, attaining a mAP@.5 of 0.5224 and an F1-score of 0.6693. Overall, the model achieved a mAP@.5 of 0.4339 and an F1-score of 0.5502.

\begin{table}[h!]
\centering
\scriptsize
\begin{tabular}{llllllllll}
\cline{1-6}
\multicolumn{6}{|c|}{Small Data 20X}     \\ \cline{1-6}
\multicolumn{1}{|l|}{}             & \multicolumn{1}{l|}{Precision}       & \multicolumn{1}{l|}{Recall}          & \multicolumn{1}{l|}{mAP@.5}           & \multicolumn{1}{l|}{mAP@.5:.95}           & \multicolumn{1}{l|}{f1-score}         \\ \cline{1-6}
\multicolumn{1}{|l|}{Bus}          & \multicolumn{1}{l|}{0.9044}          & \multicolumn{1}{l|}{0.5312}          & \multicolumn{1}{l|}{0.5224}          & \multicolumn{1}{l|}{0.3879}           & \multicolumn{1}{l|}{0.6693}           \\ \cline{1-6}
\multicolumn{1}{|l|}{Bike}         & \multicolumn{1}{l|}{0.7079}          & \multicolumn{1}{l|}{0.5193}          & \multicolumn{1}{l|}{0.4778}          & \multicolumn{1}{l|}{0.2232}           & \multicolumn{1}{l|}{0.5991}          \\ \cline{1-6}
\multicolumn{1}{|l|}{Car}          & \multicolumn{1}{l|}{0.7719}          & \multicolumn{1}{l|}{0.6903}          & \multicolumn{1}{l|}{0.6623}          & \multicolumn{1}{l|}{0.40949}          & \multicolumn{1}{l|}{0.7288}         \\ \cline{1-6}
\multicolumn{1}{|l|}{Pedestrian}   & \multicolumn{1}{l|}{0.6705}          & \multicolumn{1}{l|}{0.2355}          & \multicolumn{1}{l|}{0.2113}          & \multicolumn{1}{l|}{0.1041}           & \multicolumn{1}{l|}{0.3486}        \\ \cline{1-6}
\multicolumn{1}{|l|}{Truck}        & \multicolumn{1}{l|}{0.4638}          & \multicolumn{1}{l|}{0.3599}          & \multicolumn{1}{l|}{0.2956}          & \multicolumn{1}{l|}{0.2319}           & \multicolumn{1}{l|}{0.4053}          \\ \cline{1-6}
\multicolumn{1}{|l|}{All} & \multicolumn{1}{l|}{0.7037} & \multicolumn{1}{l|}{0.4672} & \multicolumn{1}{l|}{0.4339} & \multicolumn{1}{l|}{0.27131} & \multicolumn{1}{l|}{0.5502}   \\ \cline{1-6}
\end{tabular}
\caption{Result of YOLOv7-e6e model on the small dataset with a 20X augmentation.}
\label{tb:smalldata20x}
\end{table}

\begin{figure}[t]
  \centering
   \includegraphics[width=1\linewidth]{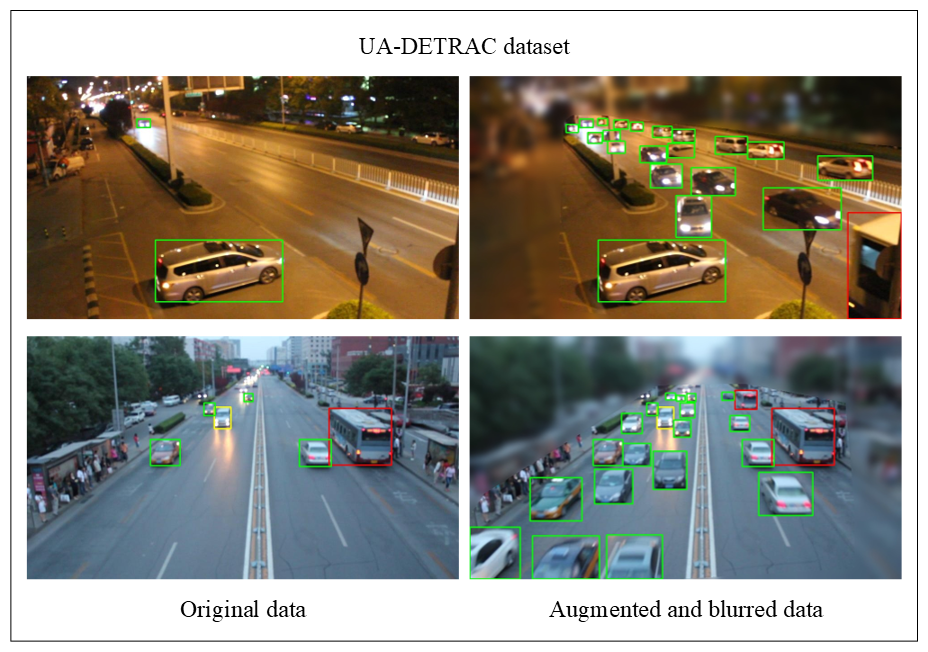}
   \caption{The comparison between original image and augmented image in UA-DETRAC dataset.}
   \label{fig:UA-DETRAC_example}
\end{figure}

Table \ref{tb:assembled} demonstrates the model's proficiency in car detection, achieving a mAP@.5 of 0.6957 and an associated F1-score of 0.7404. However, its performance was less than pedestrian detection, recording its lowest mAP@.5 at 0.176, with an F1-score of 0.3009. Furthermore, the model showed a great performance in detecting buses, attaining a mAP@.5 of 0.5692 and an F1-score of 0.6625. Overall, the model achieved a mAP@.5 of 0.5025 and an F1-score of 0.6005.

\begin{table}[h!]
\centering
\scriptsize
\begin{tabular}{llllllllll}
\cline{1-6}
\multicolumn{6}{|c|}{Assembled}  \\ \cline{1-6}
\multicolumn{1}{|l|}{}             & \multicolumn{1}{l|}{Precision}       & \multicolumn{1}{l|}{Recall}                                  & \multicolumn{1}{l|}{mAP@.5}                                   & \multicolumn{1}{l|}{mAP@.5:.95}                                  & \multicolumn{1}{l|}{f1-score}     \\ \cline{1-6}
\multicolumn{1}{|l|}{Bus}          & \multicolumn{1}{l|}{0.7274}          & \multicolumn{1}{l|}{0.6082}                                  & \multicolumn{1}{l|}{0.5692}                                  & \multicolumn{1}{l|}{0.409}                                   & \multicolumn{1}{l|}{0.6625}      \\ \cline{1-6}
\multicolumn{1}{|l|}{Bike}         & \multicolumn{1}{l|}{0.6821}          & \multicolumn{1}{l|}{0.5246}                                  & \multicolumn{1}{l|}{0.4783}                                  & \multicolumn{1}{l|}{0.225}                                   & \multicolumn{1}{l|}{0.593}      \\ \cline{1-6}
\multicolumn{1}{|l|}{Car}          & \multicolumn{1}{l|}{0.7631}          & \multicolumn{1}{l|}{0.719}                                   & \multicolumn{1}{l|}{0.6957}                                  & \multicolumn{1}{l|}{0.4393}                                  & \multicolumn{1}{l|}{0.7404}  \\ \cline{1-6}
\multicolumn{1}{|l|}{Pedestrian}   & \multicolumn{1}{l|}{0.8883}          & \multicolumn{1}{l|}{0.1812}                                  & \multicolumn{1}{l|}{0.176}                                   & \multicolumn{1}{l|}{0.0916}                                  & \multicolumn{1}{l|}{0.3009}                    \\ \cline{1-6}
\multicolumn{1}{|l|}{Truck}        & \multicolumn{1}{l|}{0.8138}          & \multicolumn{1}{l|}{0.623}                                   & \multicolumn{1}{l|}{0.5932}                                  & \multicolumn{1}{l|}{0.4038}                                  & \multicolumn{1}{l|}{0.7054}  \\ \cline{1-6}
\multicolumn{1}{|l|}{All} & \multicolumn{1}{l|}{0.7749} & \multicolumn{1}{l|}{0.5312} & \multicolumn{1}{l|}{0.5025} & \multicolumn{1}{l|}{0.3138} & \multicolumn{1}{l|}{0.6005}       \\ \cline{1-6}

\end{tabular}
\caption{Result of YOLOv7-e6e model on the assembled dataset.}
\label{tb:assembled}
\end{table}

The results of our experiments across five variations of small data are summarized in Table \ref{tb:all_results}. The highest recall, mAP@.5, and mAP@.5:.95 are 0.5312, 0.5025, and 0.3138, respectively, achieved in the assembled dataset. The highest precision and F1-score are 0.8132 and 0.62, respectively.

\begin{table}[h!]
\tiny
\centering
\begin{tabular}{|l|c|c|c|c|c|}\hline
                          & \multicolumn{1}{c|}{Precision} & \multicolumn{1}{c|}{Recall}    & \multicolumn{1}{c|}{mAP@.5}     & \multicolumn{1}{c|}{mAP@.5:.95}    & \multicolumn{1}{c|}{f1-score} \\ \hline
Small Data (SD)     &\textbf{ 0.8132} & 0.5104                         & 0.4798                         & 0.29                           & \textbf{0.62}  \\ \hline
SD 3X  & 0.8028                         & 0.4806                         & 0.4651                         & 0.3054                         & 0.5859                        \\ \hline
SD 10X & 0.7551                         & 0.5234                         & 0.4931                         & 0.2966                         & 0.6141                        \\ \hline
SD 20X & 0.7037                         & 0.4672                         & 0.4339                         & 0.27131                        & 0.5502                        \\ \hline
Assembled     & 0.7749                         & \textbf{0.5312} & \textbf{0.5025} & \textbf{0.3138} & 0.6005                        \\ \hline
\end{tabular}
\caption{Results of the YOLOv7-E6E model on multiple small datasets.}
\label{tb:all_results}
\end{table}

We observe significant improvements, with the mAP@.5 increasing from 0.4798 to 0.5025, and the mAP@.5:.95 rising from 0.29 to 0.3138  using the Assembled dataset. 

\subsection{UA-DETRAC}

\begin{table}[h!]
\scriptsize
\centering
\begin{tabular}{|llllll|}
\hline
\multicolumn{6}{|c|}{Small Data}                                                                                                                                                                  \\ \hline
\multicolumn{1}{|l|}{}             & \multicolumn{1}{l|}{Precision} & \multicolumn{1}{l|}{Recall} & \multicolumn{1}{l|}{mAP@.5} & \multicolumn{1}{l|}{ mAP@.5:.95} & f1-score \\ \hline
\multicolumn{1}{|l|}{Others}       & \multicolumn{1}{l|}{0}         & \multicolumn{1}{l|}{0}      & \multicolumn{1}{l|}{0}      & \multicolumn{1}{l|}{0}                                       & 0        \\ \hline
\multicolumn{1}{|l|}{Car}          & \multicolumn{1}{l|}{0.7135}    & \multicolumn{1}{l|}{0.7565} & \multicolumn{1}{l|}{0.7171} & \multicolumn{1}{l|}{0.5225}                                  & 0.7344   \\ \hline
\multicolumn{1}{|l|}{Van}          & \multicolumn{1}{l|}{0.6155}    & \multicolumn{1}{l|}{0.4754} & \multicolumn{1}{l|}{0.4179} & \multicolumn{1}{l|}{0.3319}                                  & 0.5365   \\ \hline
\multicolumn{1}{|l|}{Bus}          & \multicolumn{1}{l|}{0.6155}    & \multicolumn{1}{l|}{0.7771} & \multicolumn{1}{l|}{0.7359} & \multicolumn{1}{l|}{0.5607}                                  & 0.7574   \\ \hline
\multicolumn{1}{|l|}{\textbf{All}} & \multicolumn{1}{l|}{0.7387}    & \multicolumn{1}{l|}{0.4677} & \multicolumn{1}{l|}{0.4677} & \multicolumn{1}{l|}{0.3538}                                  & 0.5071   \\ \hline
\end{tabular}
\caption{Results of YOLOv7-E6E model on the small data.}
\label{tb:no_blur}
\end{table}

Figure \ref{fig:UA-DETRAC_example} shows the original and augmented samples in UA-DETRAC dataset \cite{ua_detrac}. The results presented in Table \ref{tb:no_blur} illustrate the performance of the YOLOv7-E6E model trained on a modest dataset without augmentation. Notably, the model achieved its highest mAP@.5 of 0.7359 for detecting Buses, while recording its lowest mAP@.5 of 0 for Others detection. It showed impressive performance in detecting Cars, with a mAP@.5 of 0.7171. Overall, the model exhibited a mAP@.5 of 0.4677 and an F1-score of 0.5071.

\begin{table}[h!]
\scriptsize
\centering
\begin{tabular}{|llllll|}
\hline
\multicolumn{6}{|c|}{Blurred Small Data}                                                                                                                                                              \\ \hline
\multicolumn{1}{|l|}{}             & \multicolumn{1}{l|}{Precision} & \multicolumn{1}{l|}{Recall} & \multicolumn{1}{l|}{mAP@.5} & \multicolumn{1}{l|}{mAP@.5:.95} & f1-score \\ \hline
\multicolumn{1}{|l|}{Others}       & \multicolumn{1}{l|}{0.9326}    & \multicolumn{1}{l|}{0.6032} & \multicolumn{1}{l|}{0.6062} & \multicolumn{1}{l|}{0.4496}                                  & 0.7326   \\ \hline
\multicolumn{1}{|l|}{Car}          & \multicolumn{1}{l|}{0.8458}    & \multicolumn{1}{l|}{0.7973} & \multicolumn{1}{l|}{0.7714} & \multicolumn{1}{l|}{0.5681}                                  & 0.8208   \\ \hline
\multicolumn{1}{|l|}{Van}          & \multicolumn{1}{l|}{0.6712}    & \multicolumn{1}{l|}{0.6474} & \multicolumn{1}{l|}{0.584}  & \multicolumn{1}{l|}{0.451}                                   & 0.6591   \\ \hline
\multicolumn{1}{|l|}{Bus}          & \multicolumn{1}{l|}{0.8807}    & \multicolumn{1}{l|}{0.8811} & \multicolumn{1}{l|}{0.8487} & \multicolumn{1}{l|}{0.6503}                                  & 0.8809   \\ \hline
\multicolumn{1}{|l|}{\textbf{All}} & \multicolumn{1}{l|}{0.8326}    & \multicolumn{1}{l|}{0.7323} & \multicolumn{1}{l|}{0.7026} & \multicolumn{1}{l|}{0.5298}                                  & 0.7734   \\ \hline
\end{tabular}
\caption{Results of YOLOv7-E6E model on the blurred small data.}
\label{tb:region_blur}
\end{table}

As shown in Table \ref{tb:region_blur}, the YOLOv7-E6E object detection model demonstrates its performance on a small dataset containing blurred regions. The model achieved a strong mean Average Precision (mAP@.5) of 0.7026, indicating a good overall ability to detect objects. However, its performance varied across different classes. Notably, it excelled at identifying Buses, achieving a top mAP@.5 of 0.8487. This suggests the model effectively learned the distinctive features of buses even when partially obscured. Conversely, Van detection were the most challenging, with a lowest mAP@.5 of 0.584. The F1-score of 0.7734 further supports this notion, suggesting a good balance between precision and recall for most classes, but potentially needing improvement for van detection.

\begin{table}[h!]
\scriptsize
\centering
\begin{tabular}{|llllll|}
\hline
\multicolumn{6}{|c|}{Blurred Small Data 3X}  \\ \hline
\multicolumn{1}{|l|}{}             & \multicolumn{1}{l|}{Precision} & \multicolumn{1}{l|}{Recall} & \multicolumn{1}{l|}{mAP@.5} & \multicolumn{1}{l|} {mAP@.5:.95} & f1-score \\ \hline
\multicolumn{1}{|l|}{Others}       & \multicolumn{1}{l|}{0.9071}    & \multicolumn{1}{l|}{0.6177} & \multicolumn{1}{l|}{0.6243} & \multicolumn{1}{l|}{0.4617}                                  & 0.7349   \\ \hline
\multicolumn{1}{|l|}{Car}          & \multicolumn{1}{l|}{0.8449}    & \multicolumn{1}{l|}{0.7734} & \multicolumn{1}{l|}{0.7717} & \multicolumn{1}{l|}{0.5657}                                  & 0.8076   \\ \hline
\multicolumn{1}{|l|}{Van}          & \multicolumn{1}{l|}{0.581}     & \multicolumn{1}{l|}{0.6639} & \multicolumn{1}{l|}{0.5939} & \multicolumn{1}{l|}{0.4578}                                  & 0.6197   \\ \hline
\multicolumn{1}{|l|}{Bus}          & \multicolumn{1}{l|}{0.8707}    & \multicolumn{1}{l|}{0.8861} & \multicolumn{1}{l|}{0.8664} & \multicolumn{1}{l|}{0.6257}                                  & 0.8783   \\ \hline
\multicolumn{1}{|l|}{\textbf{All}} & \multicolumn{1}{l|}{0.8009}    & \multicolumn{1}{l|}{0.7353} & \multicolumn{1}{l|}{0.7141} & \multicolumn{1}{l|}{0.5277}                                  & 0.7601   \\ \hline
\end{tabular}
\caption{Results of YOLOv7-E6E model on the blurred small data with 3X augmentation.}
\label{tb:3aug_detrac}
\end{table}

The results presented in Table \ref{tb:3aug_detrac} illustrate the performance of the YOLOv7-E6E model trained on a small dataset with 3X augmentation. The model excelled at identifying Buses, achieving a top mAP@.5 of 0.8664. Conversely, Van detection proved the most challenging, with the lowest mAP@.5 of 0.5939. This difference of 0.2725 in mAP@.5 highlights a significant performance gap. Vans may require more diverse details for accurate identification, details that might have been limited even with augmentation. The F1-score of 0.7601 further supports this notion, suggesting a good balance between precision and recall for most classes, but potentially needing improvement for van detection.

\begin{table}[h!]
\scriptsize
\centering
\begin{tabular}{|llllll|}
\hline
\multicolumn{6}{|c|}{Blurred Small Data 10X} \\ \hline
\multicolumn{1}{|l|}{}             & \multicolumn{1}{l|}{Precision} & \multicolumn{1}{l|}{Recall} & \multicolumn{1}{l|}{mAP@.5} & \multicolumn{1}{l|}{mAP@.5:.95} & f1-score \\ \hline
\multicolumn{1}{|l|}{Others}       & \multicolumn{1}{l|}{0.8873}    & \multicolumn{1}{l|}{0.6311} & \multicolumn{1}{l|}{0.6262} & \multicolumn{1}{l|}{0.4522}                                  & 0.7376   \\ \hline
\multicolumn{1}{|l|}{Car}          & \multicolumn{1}{l|}{0.7358}    & \multicolumn{1}{l|}{0.8754} & \multicolumn{1}{l|}{0.8108} & \multicolumn{1}{l|}{0.6006}                                  & 0.7995   \\ \hline
\multicolumn{1}{|l|}{Van}          & \multicolumn{1}{l|}{0.6702}    & \multicolumn{1}{l|}{0.6518} & \multicolumn{1}{l|}{0.5771} & \multicolumn{1}{l|}{0.4491}                                  & 0.6609   \\ \hline
\multicolumn{1}{|l|}{Bus}          & \multicolumn{1}{l|}{0.8615}    & \multicolumn{1}{l|}{0.9104} & \multicolumn{1}{l|}{0.8453} & \multicolumn{1}{l|}{0.6359}                                  & 0.8853   \\ \hline
\multicolumn{1}{|l|}{\textbf{All}} & \multicolumn{1}{l|}{0.7887}    & \multicolumn{1}{l|}{0.7672} & \multicolumn{1}{l|}{0.7149} & \multicolumn{1}{l|}{0.5345}                                  & 0.7708   \\ \hline
\end{tabular}
\caption{Results of YOLOv7-E6E model on the blurred small data with 10X augmentation.}
\label{tb:10aug_detrac}
\end{table}

As depicted in Table \ref{tb:10aug_detrac}, the YOLOv7-E6E object detection model showcases its potential on a limited dataset significantly enriched with various transformations (10X augmentation). This extensive augmentation approach likely furnished the model with a wider spectrum of object appearances to learn from, potentially enhancing its generalizability. The model attained a commendable overall mean Average Precision (mAP@.5) of 0.7149, indicating its proficient ability to detect objects despite the constraints of a modest dataset.

Delving deeper into the findings, we notice some fluctuations in performance across different object categories. The model excelled in identifying Buses, achieving a top mAP@.5 of 0.8453. This implies that the model effectively grasped the distinctive features of buses even with the significant dataset augmentation. Conversely, Van detection posed as the most challenging task, with the lowest mAP@.5 of 0.5771. This considerable difference of 0.2682 in mAP@.5 underscores a substantial performance gap. It's plausible that vans exhibit a more varied range of visual characteristics crucial for precise identification, details that might still be limited even with a 10X augmentation. The F1-score of 0.7708 aligns with this notion, indicating a good balance between precision and recall for most categories, albeit potentially warranting enhancement for van detection.

\begin{table}[h!]
\scriptsize
\centering
\begin{tabular}{|llllll|}
\hline
\multicolumn{6}{|c|}{Blurred Small Data 20X} \\ \hline
\multicolumn{1}{|l|}{}             & \multicolumn{1}{l|}{Precision} & \multicolumn{1}{l|}{Recall} & \multicolumn{1}{l|}{mAP@.5} & \multicolumn{1}{l|}{mAP@.5:.95} & f1-score \\ \hline
\multicolumn{1}{|l|}{Others}       & \multicolumn{1}{l|}{0.7984}    & \multicolumn{1}{l|}{0.2233} & \multicolumn{1}{l|}{0.2158} & \multicolumn{1}{l|}{0.1781}                                  & 0.349    \\ \hline
\multicolumn{1}{|l|}{Car}          & \multicolumn{1}{l|}{0.7748}    & \multicolumn{1}{l|}{0.8472} & \multicolumn{1}{l|}{0.8065} & \multicolumn{1}{l|}{0.6071}                                  & 0.8084   \\ \hline
\multicolumn{1}{|l|}{Van}          & \multicolumn{1}{l|}{0.6847}    & \multicolumn{1}{l|}{0.6461} & \multicolumn{1}{l|}{0.5736} & \multicolumn{1}{l|}{0.4507}                                  & 0.6648   \\ \hline
\multicolumn{1}{|l|}{Bus}          & \multicolumn{1}{l|}{0.8684}    & \multicolumn{1}{l|}{0.877}  & \multicolumn{1}{l|}{0.8492} & \multicolumn{1}{l|}{0.6734}                                  & 0.8727   \\ \hline
\multicolumn{1}{|l|}{\textbf{All}} & \multicolumn{1}{l|}{0.7816}    & \multicolumn{1}{l|}{0.6484} & \multicolumn{1}{l|}{0.6113} & \multicolumn{1}{l|}{0.4773}                                  & 0.674    \\ \hline
\end{tabular}
\caption{Results of YOLOv7-E6E model on the blurred small data with 20X augmentation.}
\label{tb:20aug_detrac}
\end{table}

The outcomes depicted in Table \ref{tb:20aug_detrac} showcase the performance of the YOLOv7-E6E model trained on a modest dataset with 20X augmentation.The model excelled at detecting Buses, achieving a top mAP@.5 of 0.8492. Conversely, the model struggled with the 'Others' category, achieving a much lower mAP@.5 of 0.2158. This substantial difference of 0.6334 in mAP@.5 highlights a significant performance gap. It's possible that the 'Others' category encompasses a highly diverse set of objects, making it more challenging for the model to learn comprehensive detection patterns, even with a large number of augmented images. The F1-score of 0.674 partially reflects this notion, suggesting an overall trade-off between precision and recall that might be improved for specific classes like 'Others'.

\begin{table}[h!]
\scriptsize
\centering
\begin{tabular}{|lccccc|}
\hline
\multicolumn{6}{|c|}{Assembled}  \\ \hline
\multicolumn{1}{|l|}{}       & \multicolumn{1}{l|}{Precision} & \multicolumn{1}{l|}{Recall} & \multicolumn{1}{l|}{mAP@.5} & \multicolumn{1}{l|}{ mAP@.5:.95} & \multicolumn{1}{l|}{f1-score} \\ \hline
\multicolumn{1}{|l|}{Others} & \multicolumn{1}{c|}{0.8996}    & \multicolumn{1}{c|}{0.6609} & \multicolumn{1}{c|}{0.6529} & \multicolumn{1}{c|}{0.4976}                                  & 0.762                         \\ \hline
\multicolumn{1}{|l|}{Car}    & \multicolumn{1}{c|}{0.8511}    & \multicolumn{1}{c|}{0.7945} & \multicolumn{1}{c|}{0.7945} & \multicolumn{1}{c|}{0.5769}                                  & 0.8218                        \\ \hline
\multicolumn{1}{|l|}{Van}    & \multicolumn{1}{c|}{0.564}     & \multicolumn{1}{c|}{0.6211} & \multicolumn{1}{c|}{0.5089} & \multicolumn{1}{c|}{0.4056}                                  & 0.5912                        \\ \hline
\multicolumn{1}{|l|}{Bus}    & \multicolumn{1}{c|}{0.8919}    & \multicolumn{1}{c|}{0.8857} & \multicolumn{1}{c|}{0.8581} & \multicolumn{1}{c|}{0.644}                                   & 0.8888                        \\ \hline
\multicolumn{1}{|l|}{All}    & \multicolumn{1}{c|}{0.8017}    & \multicolumn{1}{c|}{0.7406} & \multicolumn{1}{c|}{0.7036} & \multicolumn{1}{c|}{0.531}                                   & 0.766                         \\ \hline
\end{tabular}
\caption{Results of YOLOv7-E6E model on the assembled data.}
\label{tb:assembled_detrac}
\end{table}

The results presented in Table \ref{tb:assembled_detrac} illustrate the performance of the YOLOv7-E6E model trained on assembled data. The model achieved mean Average Precision (mAP@.5) of 0.7036, indicating its effectiveness in detecting objects within this varied dataset. The model demonstrated exceptional proficiency in identifying buses, attaining a peak mAP@.5 of 0.8581. Conversely, detecting vans posed the most formidable challenge, yielding the lowest mAP@.5 of 0.5089. This notable discrepancy of 0.3492 in mAP@.5 underscores a substantial performance contrast. 

\begin{table}[h!]
\tiny
\centering
\begin{tabular}{|l|c|c|c|c|c|}
\hline
\multicolumn{1}{|c|}{} & Precision       & Recall          & mAP@.5          & mAP@.5:.95 & f1-score        \\ \hline
Small Data (SD)        & 0.7387          & 0.4677          & 0.4677          & 0.3538                                  & 0.5071          \\ \hline
Blurred SD                & \textbf{0.8326} & 0.7323          & 0.7026          & 0.5298                                  & \textbf{0.7734} \\ \hline
Blurred SD 3X             & 0.8009          & 0.7353          & 0.7141          & 0.5277                                  & 0.7601          \\ \hline
Blurred SD 10X            & 0.7887          & \textbf{0.7672} & \textbf{0.7149} & \textbf{0.5345}                         & 0.7708          \\ \hline
Blurred SD 10X             & 0.7816          & 0.6484          & 0.6113          & 0.4773                                  & 0.674           \\ \hline
Assembled              & 0.8017          & 0.7406          & 0.7036          & 0.531                                   & 0.766           \\ \hline
\end{tabular}
\caption{Results of the YOLOv7-E6E model on multiple small datasets.}
\label{tb:all_results_detrac}
\end{table}

Our experiments exploring the YOLOv7-E6E model's performance on various small datasets with augmentation techniques are summarized in Table \ref{tb:all_results_detrac}. While 10X augmentation achieved the highest overall mAP@.5 of 0.7149 indicating balanced performance, small data with blurred images yielded the highest precision of 0.8326 and F1-score of 0.7734, suggesting improved detection accuracy. 

\section{Conclusion}

In this work, we proposed a simple augmentation technique for the object detection dataset, which is suitable for stationary camera-based datasets. Through extensive testing on two traffic monitoring datasets, we demonstrate that our augmentation approach can enhance model performance, particularly in scenarios where only a limited number of labeled samples are available and a number of samples per class is imbalanced. It achieved the performance level of the model trained on the entire dataset while utilizing only 8.5 percent of the original dataset. However, our suggested augmentation method is only tailored for stationary camera surveillance scenarios, restricting its scalability for general object detection data augmentations. Moreover, in-place augmentation is done by copy-pasting the region of object bounding boxes, which might also include irrelevant image parts from the original sample. Therefore, segmentation labels could be more appropriate, and further investigation is needed. 

\section{Acknowledgement}
This research has been supported by the Emirates Center for Mobility Research (ECMR) through Grant 12R012, United Arab Emirates University (UAEU), United Arab Emirates.

{
    \small
    \bibliographystyle{ieeenat_fullname}
    \bibliography{main.bib}
}

\end{document}